\documentclass[runningheads]{llncs}
\usepackage{graphicx}
\usepackage{amsmath,amssymb} 
\usepackage{color}
\usepackage{booktabs}
\usepackage[width=122mm,left=12mm,paperwidth=146mm,height=193mm,top=12mm,paperheight=217mm]{geometry}
\usepackage{multirow}
\usepackage{subfigure}
\usepackage{xcolor,colortbl}

\newcommand{\comment}[1]{}

\newcommand{\brho}{\mathbf{\rho}}
\newcommand{\bmu}{\mathbf{\mu}}
\newcommand{\bphi}{\mathbf{\phi}}

\newcommand{\bb}{\mathbf{b}}

\newcommand{\bp}{\mathbf{p}}
\newcommand{\bP}{\mathbf{P}}
\newcommand{\bd}{\mathbf{d}}

\newcommand{\bS}{\mathbf{S}}
\newcommand{\bW}{\mathbf{W}}

\newcommand{\bx}[0]{\mathbf{x}}
\newcommand{\by}[0]{\mathbf{y}}

\newcommand{\softargmax}{\text{softargmax}}
\newcommand{\softmax}{\text{softmax}}
\newcommand{\Rot}{\text{Rot}}
\newcommand{\Crop}{\text{Crop}}
\newcommand{\RotCrop}{\text{Rot}}

\newcommand{\desc}{\text{desc}}
\newcommand{\loss}{\mathcal{L}}

\definecolor{orange}{rgb}{1,0.5,0}
\definecolor{blue}{rgb}{0,0,0.6}

\definecolor{color1}{RGB}{0,199,1}
\definecolor{color2}{RGB}{224,43,28}

\newcommand{\kmyi}[1]{#1}
\newcommand{\kmyirmk}[1]{}

\newcommand{\pfrmk}[1]{}

\definecolor{darkolivegreen}{rgb}{0.5, 0.7, 0.3}
\newcommand{\vincentrmk}[1]{}

\newcommand{\eduard}[1]{#1}
\newcommand{\eduardrmk}[1]{}

\begin{document}
\pagestyle{headings}
\mainmatter
\def\ECCV16SubNumber{1377}  

\title{LIFT: Learned Invariant Feature Transform} 

\titlerunning{LIFT: Learned Invariant Feature Transform}

\authorrunning{K.~M. Yi, E. Trulls, V. Lepetit, P. Fua}

\author{Kwang Moo Yi$^{*,1}$, Eduard Trulls$^{*,1}$, Vincent Lepetit$^{2}$, Pascal Fua$^{1}$}

\institute{%
$^1$Computer Vision Laboratory, Ecole Polytechnique F\'{e}d\'{e}rale de Lausanne (EPFL)\\
$^2$Institute for Computer Graphics and Vision, Graz University of Technology\\
\email{\{kwang.yi, eduard.trulls, pascal.fua\}@epfl.ch, lepetit@icg.tugraz.at}%
}

\maketitle

\makeatletter
\def\blfootnote{\gdef\@thefnmark{}\@footnotetext}
\makeatother
\blfootnote{$*$ First two authors contributed equally.}
\blfootnote{This work was supported in part by the EU FP7 project MAGELLAN under grant number ICT-FP7-611526.}


\begin{abstract}

We  introduce a  novel Deep  Network  architecture that  implements the  full
feature point handling pipeline,  that is, detection, orientation estimation,
and feature description. While previous  works have successfully tackled each
one of these problems individually, we show how to learn to do all three in a
unified  manner  while  preserving  end-to-end  differentiability.   We  then
demonstrate that our Deep pipeline  outperforms state-of-the-art methods on a
number of benchmark datasets, without the need of retraining.


\end{abstract}


\section{Introduction}

Local features  play a key role  in many Computer Vision  applications. Finding
and  matching them  across  images has  been  the subject  of  vast amounts  of
research. Until recently, the best  techniques relied on carefully hand-crafted
features~\cite{Lowe04,Bay08,Tola08,Rublee11,Mainali14}.   Over  the   past  few
years, as in many areas of  Computer Vision, methods based in Machine Learning,
and  more  specifically  Deep  Learning,   have  started  to  outperform  these
traditional methods~\cite{Verdie15,Han15,Zagoruyko15,Yi16,Simo-Serra15}.

These  new algorithms,  however, address  only a  single step  in the  complete
processing  chain,  which  includes  detecting the  features,  computing  their
orientation, and extracting robust representations  that allow us to match them
across images.   In this paper we  introduce a novel Deep architecture
that performs all three steps together.  We demonstrate that it achieves better
overall performance than the state-of-the-art methods, in large part because it
allows these  individual steps to be  optimized to perform well  in conjunction
with each other.

\begin{figure}[!t]
	\centering
	\includegraphics[width=\linewidth]{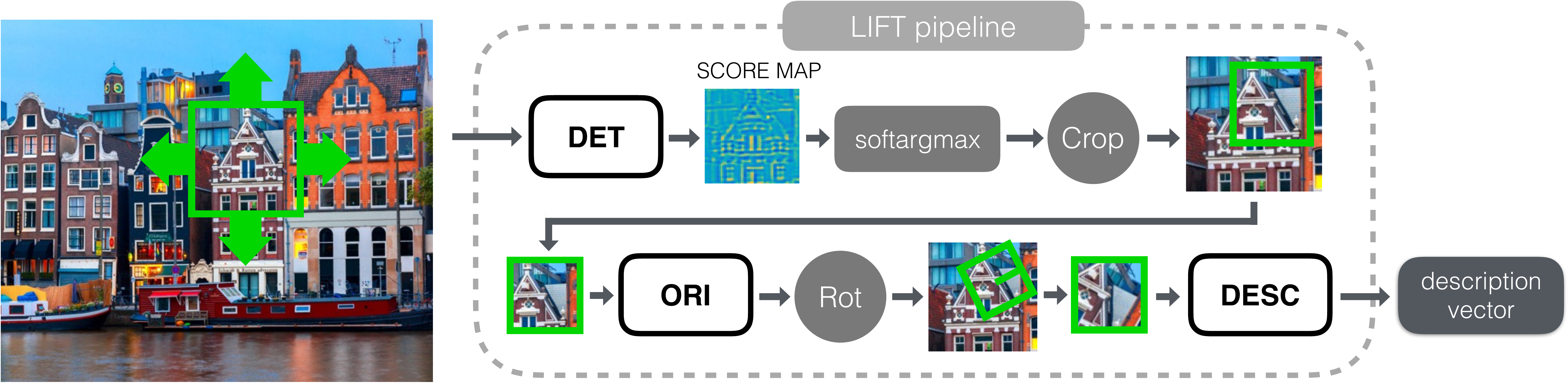}
	\caption{Our  integrated  feature  extraction pipeline.   Our  pipeline
          consists  of three  major components:  the Detector,  the Orientation
          Estimator, and the Descriptor. They are tied together
          with    differentiable    operations     to    preserve    end-to-end
          differentiability.\protect\footnotemark}
	\label{fig:architecture}
\end{figure}

Our  architecture, which  we refer  to as  LIFT for  Learned Invariant  Feature
Transform, is  depicted by  Fig.~\ref{fig:architecture}.  It consists  of three
components that feed into  each other: the Detector,
the  Orientation  Estimator,  and  the   Descriptor.   Each  one  is  based  on
Convolutional    Neural   Networks~(CNNs),    and   patterned    after   recent
ones~\cite{Verdie15,Yi16,Simo-Serra15} that  have been  shown to  perform these
individual functions  well.  To mesh  them together we use  Spatial
Transformers~\cite{Jaderberg15}
to rectify the image  patches given the output of the
Detector  and  the  Orientation  Estimator. We  also  replace  the  traditional
approaches  to   non-local  maximum  suppression   (NMS)  by  the   soft  argmax
function~\cite{Chapelle2009}.    This   allows   us  to   preserve   end-to-end
differentiability, and results in a full network that can still be trained with
back-propagation, which is not the case of any other architecture we know of.

Also, we  show {\em how} to  learn such a  pipeline in an effective  manner. To
this end,  we build  a Siamese network  and train it  using the  feature points
produced by a Structure-from-Motion (SfM) algorithm  that we ran on images of a
scene captured under different viewpoints and lighting conditions, to learn its
weights.   We formulate  this training  problem on  image patches  extracted at
different scales to  make the optimization tractable. In practice,  we found it
impossible to train the full  architecture from scratch, because the individual
components try to  optimize for different objectives.  Instead,  we introduce a
problem-specific  learning  approach to  overcome  this  problem.  It  involves
training the  Descriptor first,  which is  then used  to train  the Orientation
Estimator, and  finally the Detector,  based on the already  learned Descriptor
and Orientation Estimator, differentiating through  the entire network. At test
time, we decouple the Detector, which runs over the whole image in scale space,
from  the  Orientation  Estimator  and   Descriptor,  which  process  only  the
keypoints.

\addtocounter{footnote}{-1}
\footnotetext{Figures are best viewed in color.}

In the next section we briefly discuss earlier approaches. We then present our
approach in detail and show that it outperforms many state-of-the-art methods.


\section{Related work}

The amount of  literature relating to local features is  immense, 
but it  always revolves  about finding  feature points,
computing  their orientation,  and matching  them.   In this  section, we  will
therefore discuss these three elements separately.

\subsection{Feature Point Detectors}

Research on  feature point detection  has focused mostly on  finding distinctive
locations  whose   scale  and   rotation  can   be  reliably   estimated.  Early
works~\cite{Harris88,Moravec80}  used first-order  approximations  of the  image
signal  to  find corner  points  in  images. FAST~\cite{Rosten06}  used  Machine
Learning  techniques but  only  to  speed up  the  process  of finding  corners.
Other  than  corner  points,  SIFT~\cite{Lowe04} detect  blobs  in  scale-space;
SURF~\cite{Bay08}  use  Haar   filters  to  speed  up   the  process;  Maximally
Stable  Extremal  Regions  (MSER)~\cite{Matas02} detect  regions;  \cite{Miko02}
detect  affine  regions. SFOP~\cite{Forstner09}  use  junctions  and blobs,  and
Edge~Foci~\cite{Zitnick11}  use edges  for robustness  to illumination  changes.
More recently, feature points based on more sophisticated and carefully designed
filter responses~\cite{Mainali14,Mainali13}  have also been proposed  to further
enhance the performance of feature point detectors.

In contrast to these approaches that focus on better engineering, and following
the   early   attempts   in   learning   detectors~\cite{Sochman07,Trujillo06},
\cite{Verdie15}  showed   that  a   detector  could   be  learned   to  deliver
significantly  better performance  than  the state-of-the-art.   In this  work,
piecewise-linear convolutional  filters are learned to  robustly detect feature
points in spite of lighting and seasonal changes.  Unfortunately, this was done
only  for a  single scale  and from  a dataset  without viewpoint  changes.  We
therefore took  our inspiration from it  but had to extend  it substantially to
incorporate it into our pipeline.

\subsection{Orientation Estimation}

Despite the fact that  it plays a critical role in  matching feature points, the
problem of estimating a discriminative  orientation has received noticeably less
attention  than  detection or  feature  description.  As  a result,  the  method
introduced by SIFT~\cite{Lowe04} remains the {\em de facto} standard up to small
improvements, such  as the fact  that it can be  sped-up by using  the intensity
centroid, as in ORB~\cite{Rublee11}.

A departure from this can be found in a recent paper~\cite{Yi16} that introduced
a Deep Learning-based approach to  predicting stable orientations. This resulted
in significant gains over the state-of-the-art. We incorporate this architecture
into our pipeline  and show how to train it  using our problem-specific training
strategy, given our learned descriptors.

\subsection{Feature Descriptors}

Feature descriptors  are designed  to provide discriminative  representations of
salient image patches,  while being robust to transformations  such as viewpoint
or  illumination  changes. The  field  reached  maturity with  the  introduction
of  SIFT~\cite{Lowe04}, which  is  computed from  local  histograms of  gradient
orientations, and  SURF~\cite{Bay08}, which uses integral  image representations
to speed up the computation.  Along similar lines, DAISY~\cite{Tola08} relies on
convolved maps of oriented gradients to approximate the histograms, which yields
large computational gains when extracting dense descriptors.

Even though they have been extremely successful, these hand-crafted descriptors
can now be outperformed by newer ones that have been learned.  These range from
unsupervised  hashing  to  supervised   learning  techniques  based  on  linear
discriminant          analysis~\cite{strecha12,Winder07},         \kmyi{genetic
  algorithm~\cite{Perez13},} and convex optimization~\cite{Simonyan14}. An even
more recent trend  is to extract features directly from  raw image patches with
CNNs  trained on  large volumes  of data.   For example,  MatchNet~\cite{Han15}
trained a Siamese CNN for feature representation, followed by a fully-connected
network to learn the  comparison metric.  DeepCompare~\cite{Zagoruyko15} showed
that  a  network  that  focuses  on  the  center  of  the  image  can  increase
performance.  The approach of~\cite{Zbontar15} relied on a similar architecture
to      obtain      state-of-the-art      results      for      narrow-baseline
stereo. In~\cite{Simo-Serra15}, hard negative mining  was used to learn compact
descriptors  that use  on the  Euclidean distance  to measure  similarity.  The
algorithm of~\cite{Balntas16} relied on sample triplets to mine hard negatives.

In this  work, we  rely on the  architecture of~\cite{Simo-Serra15}  because the
corresponding descriptors are trained and  compared with the Euclidean distance,
which has a wider range of applicability than descriptors that require a learned
metric.


\section{Method}
\label{sec:method}

\begin{figure}[!t]
	\centering
	\includegraphics[width=0.9\linewidth]{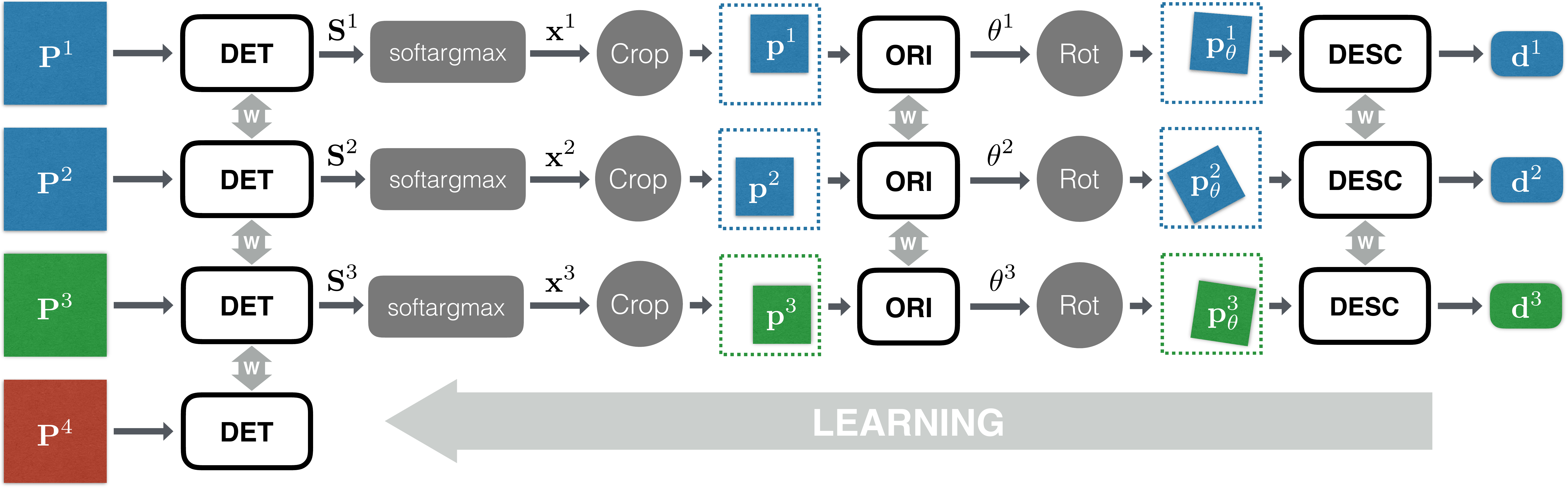}
        \caption{Our Siamese  training architecture  with four  branches, which
          takes as input  a quadruplet of patches: Patches  $\bP^1$ and $\bP^2$
          (blue) correspond to different views  of the same physical point, and
          are  used  as positive  examples  to  train the  Descriptor;  $\bP^3$
          (green)  shows a  different  3D  point, which  serves  as a  negative
          example for the Descriptor; and $\bP^4$ (red) contains no distinctive
          feature points  and is only used  as a negative example  to train the
          Detector. Given a  patch $\bP$, the Detector,  the $\softargmax$, and
          the Spatial Transformer layer $\Crop$  provide all together a smaller
          patch  $\bp$ inside  $\bP$.  $\bp$  is then  fed  to the  Orientation
          Estimator,  which along  with the  Spatial Transformer  layer $\Rot$,
          provides the  rotated patch $\bp_{\theta}$  that is processed  by the
          Descriptor to obtain the final description vector $\bd$.}
	\label{fig:training}
\end{figure}

In  this  section,  we  first   formulate  the  entire  feature  detection  and
description  pipeline  in  terms  of   the  Siamese  architecture  depicted  by
Fig.~\ref{fig:training}.
Next, we  discuss the  type of data  we need  to train our  networks and  how to
collect it. We then describe the training procedure in detail.

\subsection{Problem formulation}

We use image patches as input, rather than full images. This makes the learning
scalable without  loss of  information, as  most image  regions do  not contain
keypoints.   The  patches are  extracted  from  the  keypoints  used by  a  SfM
pipeline, as will be discussed in Section~\ref{subsec:dataset}.
We  take them  to be  small enough  that  we can  assume they  contain only  one
dominant local feature at the given scale, which reduces the learning process to
finding the most distinctive point in the patch.

To train our network we create  the four-branch Siamese architecture pictured in
Fig.~\ref{fig:training}. Each  branch contains three distinct  CNNs, a Detector,
an  Orientation Estimator,  and  a  Descriptor. For  training  purposes, we  use
quadruplets of  image patches. Each one  includes two image patches  $\bP^1$ and
$\bP^2$, that  correspond to  different views  of the same  3D point,  one image
patch $\bP^3$,  that contains the  projection of a  different 3D point,  and one
image patch $\bP^4$ that does not  contain any distinctive feature point. During
training, the $i$-th patch $\bP^i$ of each quadruplet will go through the $i$-th
branch.

To  achieve end-to-end  differentiability,  the components  of  each branch  are
connected as follows:
\begin{enumerate}
\item Given an input image patch $\bP$, the Detector provides a score map $\bS$.
\item We  perform a soft argmax~\cite{Chapelle2009}  on the score map  $\bS$ and
return the location $\bx$ of a single potential feature point.
\item  We extract  a  smaller patch  $\bp$  centered on  $\bx$  with the Spatial
Transformer layer $\Crop$ (Fig.~\ref{fig:training}). This serves as
the input to the Orientation Estimator.
\item The Orientation Estimator predicts a patch orientation $\theta$.
\item  We rotate  $\bp$ according  to this  orientation using  a second  Spatial
Transformer layer, labeled as  $\RotCrop$ in Fig.~\ref{fig:training}, to produce
$\bp_\theta$.
\item $\bp_\theta$ is fed to the Descriptor network, which
computes a feature vector $\bd$.
\end{enumerate}

Note that the  Spatial Transformer layers are used only  to manipulate the image
patches while preserving differentiability. They  are not learned modules. Also,
both the  location $\bx$ proposed by  the Detector and the  orientation $\theta$
for the  patch proposal are treated  implicitly, meaning that we  let the entire
network discover distinctive locations and stable orientations while learning.

Since our network  consists of components with different  purposes, learning the
weights is  non-trivial. Our early attempts  at training the network  as a whole
from  scratch  were  unsuccessful.  We  therefore  designed  a  problem-specific
learning  approach  that  involves  learning  first  the  Descriptor,  then  the
Orientation Estimator  given the learned  descriptor, and finally  the Detector,
conditioned on the  other two. This allows us to  tune the Orientation Estimator
for the Descriptor, and the Detector for the other two components.

We will elaborate on  this learning strategy in Secs.~\ref{subsec:descriptor}
(Descriptor),    \ref{subsec:orientation}     (Orientation    Estimator),    and
\ref{subsec:detector} (Detector), that is, in the order they are learned.

\subsection{Creating the Training Dataset}
\label{subsec:dataset}

There    are    datasets    that    can    be    used    to    train    feature
descriptors~\cite{Winder07} and orientation  estimators~\cite{Yi16}. However it
is not  so clear how  to train  a keypoint detector,  and the vast  majority of
techniques    still    rely    on    hand-crafted    features.     The    TILDE
detector~\cite{Verdie15} is  an exception,  but the  training dataset  does not
exhibit any viewpoint changes.

\newcommand{\ima}{2.8cm}
\begin{figure}[!tp]
  \centering
  \includegraphics[height=\ima]{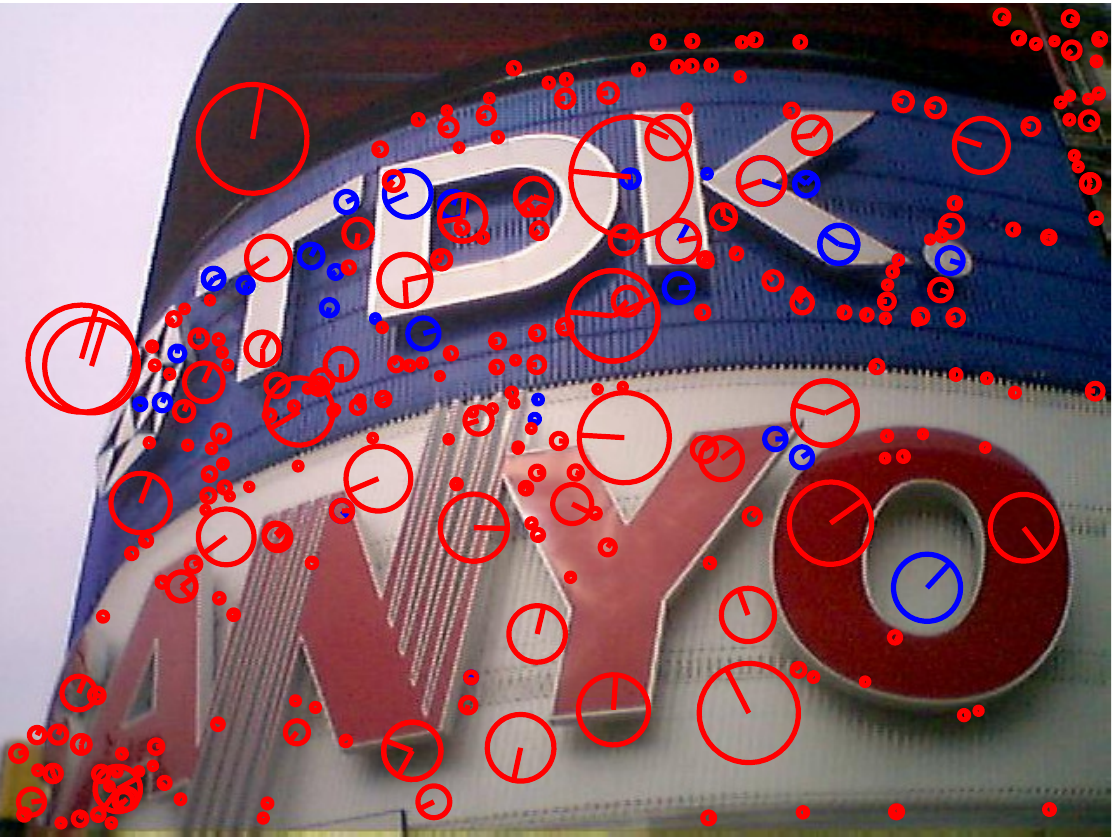}\hspace{0.5mm}%
  \includegraphics[height=\ima]{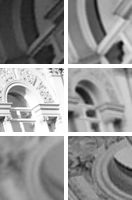}\hspace{5mm}%
  \includegraphics[height=\ima]{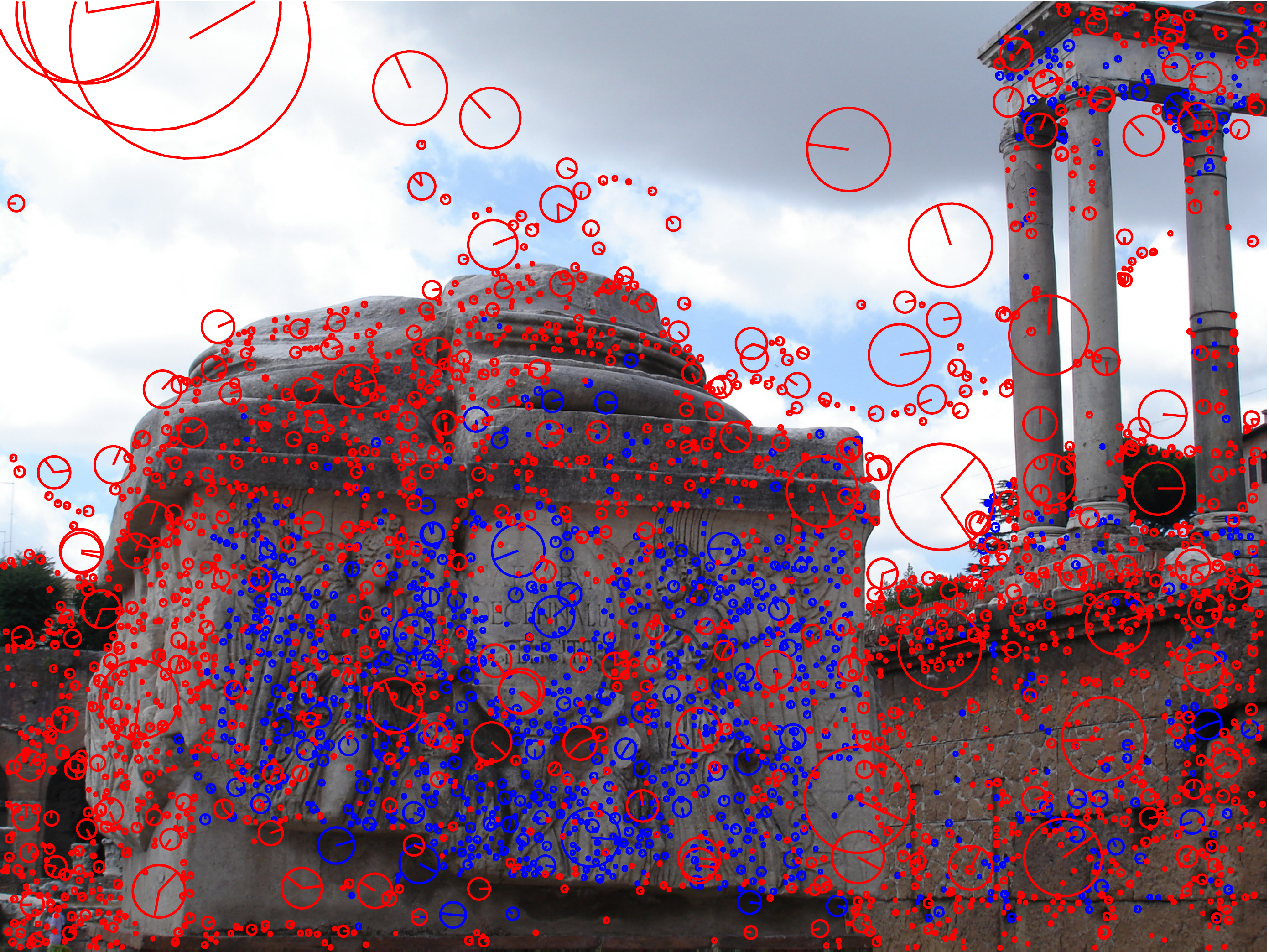}\hspace{0.5mm}%
  \includegraphics[height=\ima]{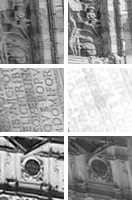}%
  \caption{Sample  images and  patches from  {\it Piccadilly}  (left) and
    {\it Roman-Forum}  (right).  Keypoints that survive  the SfM pipeline
    are drawn in blue, and the rest in red.}
  \label{fig:datasets}
\end{figure}

To achieve invariance we need images that capture views of the same scene under
different illumination conditions and seen from different perspectives. We thus
turned to  photo-tourism image sets.   We used the collections  from Piccadilly
Circus  in  London  and  the  Roman  Forum  in  Rome  from  \cite{Wilson14}  to
reconstruct the 3D using VisualSFM~\cite{Wu13},  which relies of SIFT features.
{\it Piccadilly}  contains 3384 images,  and the reconstruction has  59k unique
points with an average of 6.5 observations for each. {\it Roman-Forum} contains
1658 images and 51k unique points, with an average of 5.2 observations for each.
Fig.~\ref{fig:datasets} shows some examples.

We  split the  data  into  training and  validation  sets,  discarding views  of
training points  on the  validation set  and vice-versa.  To build  the positive
training  samples we  consider  only the  feature points  that  survive the  SfM
reconstruction process. To  extract patches that do not  contain any distinctive
feature point,  as required  by our  training method,  we randomly  sample image
regions that  contain no SIFT  features, including those  that were not  used by
SfM.

We extract  grayscale training patches  according to  the scale $\sigma$  of the
point, for both  feature and non-feature point image regions.  Patches $\bP$ are
extracted from a  $24\sigma \times 24\sigma$ support region  at these locations,
and standardized  into $S \times  S$ pixels  where $S=128$. The  smaller patches
$\bp$ and $\bp_\theta$ that serve as  input to the Orientation Estimator and the
Descriptor, are cropped and rotated versions  of these patches, each having size
$s \times  s$, where $s=64$. The  smaller patches effectively correspond  to the
SIFT descriptor support region size of $12\sigma$. To avoid biasing the data, we
apply uniform random perturbations to the  patch location with a range of $20\%$
(4.8$\sigma$). Finally,  we normalize  the patches with  the grayscale  mean and
standard deviation of the entire training set.

\subsection{Descriptor}
\label{subsec:descriptor}

Learning  feature  descriptors from  raw  image  patches has  been  extensively
researched                   during                  the                   past
year~\cite{Han15,Zagoruyko15,Simo-Serra15,Zbontar15,Balntas16,Paulin15},   with
multiple works reporting impressive results on patch retrieval, narrow baseline
stereo, and  matching non-rigid deformations.   Here we rely on  the relatively
simple  networks   of~\cite{Simo-Serra15},  with  three   convolutional  layers
followed by hyperbolic tangent units, $l_2$ pooling~\cite{Sermanet12} and local
subtractive normalization, as they do not require learning a metric.

The \eduard{Descriptor} can be formalized simply as
\begin{equation}
\label{eq:descriptor}
\bd = h_\rho(\bp_\theta) \> ,
\end{equation}
where $h(.)$  denotes the \eduard{Descriptor}  CNN, $\brho$ its  parameters, and
$\bp_\theta$ is the rotated patch  from the Orientation Estimator. When training
the Descriptor,  we do not yet  have the Detector and  the Orientation Estimator
trained. We  therefore use the image  locations and orientations of  the feature
points used by the SfM to generate image patches $\bp_\theta$.

\eduard{We train the Descriptor  by minimizing the sum of the  loss for pairs of
corresponding patches $\left(\bp_\theta^1, \bp_\theta^2\right)$ and the loss for
pairs  of non-corresponding  patches $\left(\bp_\theta^1,  \bp_\theta^3\right)$.
The loss  for pair $\left(\bp_\theta^k,  \bp_\theta^l\right)$ is defined  as the
hinge  embedding  loss  of  the Euclidean  distance  between  their  description
vectors. We write
\begin{equation}
  \label{eq:descriptor_loss}
	\loss_\desc(\bp_\theta^k,  \bp_\theta^l) = 
\begin{cases}
  \left\| 
    h_\rho\!\left(\bp_\theta^k\right) - h_\rho\!\left(\bp_\theta^l\right) 
  \right\|_2 & \text{for positive pairs, and} \\
  \max\left(
    0, C - \left\| h_\rho\!\left(\bp_\theta^k\right) - h_\rho\!\left(\bp_\theta^l\right) \right\|_2
  \right) & \text{for negative pairs} \> ,
\end{cases}
\end{equation}
}%
where positive  and negative  samples are  pairs of  patches that  do or  do not
correspond to  the same physical  3D points,  $\left\| \cdot \right\|_2$  is the
Euclidean distance, and $C=4$ is the margin for embedding.

We use hard  mining during training, which was shown  in ~\cite{Simo-Serra15} to
be critical for  descriptor performance. Following this  methodology, we forward
$K_f$ sample pairs and  use only the $K_b$ pairs with  the highest training loss
for  back-propagation, where  $r =  K_f /  K_b \geq  1$ is  the `mining  ratio'.
In~\cite{Simo-Serra15}  the  network was  pre-trained  without  mining and  then
fine-tuned with $r=8$.  Here, we use an increasing mining  scheme where we start
with  $r=1$ and  double the  mining ratio  every 5000  batches. We  use balanced
batches with 128 positive pairs and 128 negative pairs, mining each separately.

\subsection{Orientation Estimator}
\label{subsec:orientation}

Our  Orientation Estimator  is inspired  by that  of~\cite{Yi16}. However,  this
specific  one  requires pre-computations  of  description  vectors for  multiple
orientations to compute  numerically the Jacobian of the  method parameters with
respect to orientations.  This is a critical limitation for  us because we treat
the output of  the detector component implicitly throughout the  pipeline and it
is thus not possible to pre-compute the description vectors.

We therefore propose to  use Spatial Transformers~\cite{Jaderberg15} instead to
learn the  orientations.  Given a patch  $\bp$ from the region  proposed by the
detector, the Orientation Estimator predicts an orientation
\begin{equation}
\theta = g_\phi(\bp) \> ,
\end{equation}
where $g$ denotes the Orientation Estimator CNN, and $\bphi$ its parameters.

Together with the location $\bx$ from the Detector and $\bP$ the original image
patch,  $\theta$  is  then  used   by  the  second  Spatial  Transformer  Layer
$\RotCrop(.)$             to             provide            a             patch
$\bp_\theta=\RotCrop\left(\bP,\bx,\theta\right)$, which is  the rotated version
of patch $\bp$.

\newcommand{\ori}{{\text{orientation}}}

We train the Orientation Estimator to provide the orientations that minimize the
distances between description vectors for different views of the same 3D points.
We use the already trained Descriptor to compute the description vectors, and as
the Detector is still not trained, we use the image locations from SfM.
\eduard{More formally, we minimize the loss for pairs of corresponding patches,
defined as the Euclidean distance between their description vectors}
\begin{equation}
  \label{eq:orientation_loss}
	\loss_\ori(\bP^1, \bx^1, \bP^2, \bx^2) = 
  \left\|
  h_\rho(G(\bP^1, \bx^1)) -
  h_\rho(G(\bP^2, \bx^2))
  \right\|_2
\> ,
\end{equation}
where $G(\bP, \bx)$ is the patch centered on $\bx$ after orientation correction:
$G(\bP,  \bx)  =  \RotCrop(\bP,  \bx, g_\phi(\Crop(\bP,  \bx)))$.  This  complex
notation is  necessary to  properly handle  the cropping  of the  image patches.
Recall  that  pairs  $(\bP^1,  \bP^2)$ comprise  image  patches  containing  the
projections of the  same 3D point, and locations $\bx^1$  and $\bx^2$ denote the
reprojections of  these 3D points. As  in~\cite{Yi16}, we do not  use pairs that
correspond to different physical points whose orientations are not related.

\subsection{Detector}
\label{subsec:detector}

The  Detector takes  an image  patch as  input, and  returns a  score map.   We
implement it  as a  convolution layer followed  by piecewise  linear activation
functions, as  in TILDE \cite{Verdie15}.   More precisely, the score  map $\bS$
for patch $\bP$ is computed as:
\begin{equation}
  \bS = f_\bmu(\bP) = \sum_n^N{\delta_n\max^M_m{\left(\bW_{mn} \ast \bP + \bb_{mn}\right)}} \>,
\end{equation}
where  $f_\bmu(\bP)$  denotes  the   Detector  itself  with  parameters  $\bmu$,
$\delta_n$ is  $+1$ if  $n$ is  odd and $-1$  otherwise, $\bmu$  is made  of the
filters $W_{mn}$ and  biases $b_{mn}$ of the convolution layer  to learn, $\ast$
denotes the \eduard{convolution operation}, and $N$ and $M$ are hyper-parameters
controlling the complexity of the piecewise linear activation function.

The main  difference with TILDE lies  in the way  we train this layer.   To let
$\bS$ have maxima  in places other than a fixed  location retrieved by SfM,
we  treat this  location  implicitly,  as a  latent  variable.   Our method  can
potentially  discover  points that  are  more  reliable  and easier  to  learn,
whereas~\cite{Verdie15}  cannot. Incidentally,  in  our  early experiments,  we
noticed that it was harmful to force  the Detector to optimize directly for SfM
locations.

\newcommand{\bby}{{\bf y}}
  
From the score map $\bS$, we obtain the location $\bx$ of a feature point as
\begin{equation}
  \bx = \softargmax \left( \bS \right) \> ,
\end{equation}
where $\softargmax$  is a function which  computes the Center of  Mass with the
weights      being     the      output      of      a     standard      softmax
function~\cite{Chapelle2009}. We write
\begin{equation}
  \softargmax \left( \bS \right) =
  \frac{\sum_\bby{\exp(\beta\bS(\bby))\bby}}
  {\sum_\bby{\exp(\beta\bS(\bby))}}
\> ,
\end{equation}
where  $\by$ are locations in $\bS$,  and $\beta=10$  is a  hyper-parameter
controlling the  smoothness of  the $\softargmax$. This  $\softargmax$ function
acts as  a differentiable version of  non-maximum suppression. $\bx$ is  given to
the first Spatial Transformer Layer $\Crop(.)$ together with the patch $\bP$ to
extract a smaller patch $\bp =  \Crop\left(\bP,\bx\right)$ used as input to the
Orientation Estimator.

As the Orientation Estimator and the Descriptor have been learned by this point,
we  can  train the  Detector  given  the full  pipeline.  To  optimize over  the
parameters $\bmu$, we minimize the distances between description vectors for the
pairs of patches  that correspond to the same physical  points, while maximizing
the  classification score  for patches  not corresponding  to the  same physical
points.

More  exactly,   given   training  quadruplets  $\left(\bP^1,   \bP^2,  \bP^3,
\bP^4\right)$, where $\bP^1$ and $\bP^2$  correspond to the same physical point,
$\bP^1$  and $\bP^3$  correspond  to  different SfM  points,  and  $\bP^4$ to  a
non-feature point location, we minimize the sum of their loss functions
\begin{equation}
  \label{eq:detector_loss}
	\loss_{\text{detector}}(\bP^1,   \bP^2,    \bP^3,   \bP^4)    =   \gamma
  \loss_{class}(\bP^1, \bP^2, \bP^3, \bP^4) + \loss_{pair}(\bP^1,   \bP^2) \> ,
\end{equation}
where $\gamma$ is a hyper-parameter balancing the two terms in this summation
\begin{equation}
  \loss_{\text{class}}(\bP^1, \bP^2, \bP^3, \bP^4)  =
  \sum_{i=1}^{4} \alpha_i\max\left(0,   \left(1    -
      \softmax\left(f_{\bmu}\left(\bP^i\right)\right)y_i\right)\right)^2
  \> ,
\end{equation}
with  $y_i=-1$ and  $\alpha_i=3/6$ if  $i=4$, and  $y_i=+1$ and  $\alpha_i=1/6$
otherwise  to  balance   the  positives  and  negatives.    $\softmax$  is  the
log-mean-exponential softmax function. We write
\begin{equation}
  \label{eq:detector_pair_loss}
\begin{array}{lllll}
  \loss_{\text{pair}}(\bP^1,   \bP^2) = \| &&
  h_\rho(G(\bP^1, \softargmax(f_\bmu(\bP^1)))) &-&\\
&&   h_\rho(G(\bP^2, \softargmax(f_\bmu(\bP^2))))
 && \|_2 \> .
\end{array}
\end{equation}

Note  that the  locations  of the  detected feature  points  $\bx$ appear  only
implicitly  and   are  discovered  during  training.   Furthermore,  all  three
components are tied  in with the Detector learning.  As  with the Descriptor we
use a hard mining strategy, in this case with a fixed mining ratio of $r=4$.

In practice,  as the Descriptor already  learns some invariance, it  can be hard
for the  Detector to find  new points to learn  implicitly. To let  the Detector
start with an idea  of the regions it should find, we  first constrain the patch
proposals $\bp  = \Crop(\bP,  \softargmax(f_\bmu(\bP)))$ that correspond  to the
same physical points to overlap. We  then continue training the Detector without
this constraint.

Specifically, when pre-training the  Detector, we replace $\loss_{\text{pair}}$
in    Eq.~(\ref{eq:detector_loss})   with    $\tilde{\loss}_{\text{pair}}$,   where
$\tilde{\loss}_{\text{pair}}$  is  equal to  0  when  the patch  proposals  overlap
exactly, and increases  with the distance between them  otherwise. We therefore
write
\begin{equation}
\label{eq:detector_pair_pre}
  \tilde{\loss}_{\text{pair}}(\bP^1,   \bP^2) = 
  1 - \frac{\bp^1 \cap \bp^2}{\bp^1 \cup\bp^2} +
  \frac{\max\left(0, \left\|\bx^1 -  \bx^2\right\|_1 - 2s\right)}{\sqrt[]{\bp^1
      \cup\bp^2}}
\> ,
\end{equation}
where  $\bx^j =  \softargmax(f_\bmu(\bP^j))$,  $\bp^j  = \Crop(\bP^j,  \bx^j)$,
$\left\|\cdot\right\|_1$ is the $l_1$ norm. Recall that $s=64$ pixels is the
width and height of the patch proposals.

\subsection{Runtime pipeline}
\label{sec:method:testing}

\begin{figure}[!t]
  \centering
  \includegraphics[width=0.8\linewidth, trim={0 0 0 0},clip]{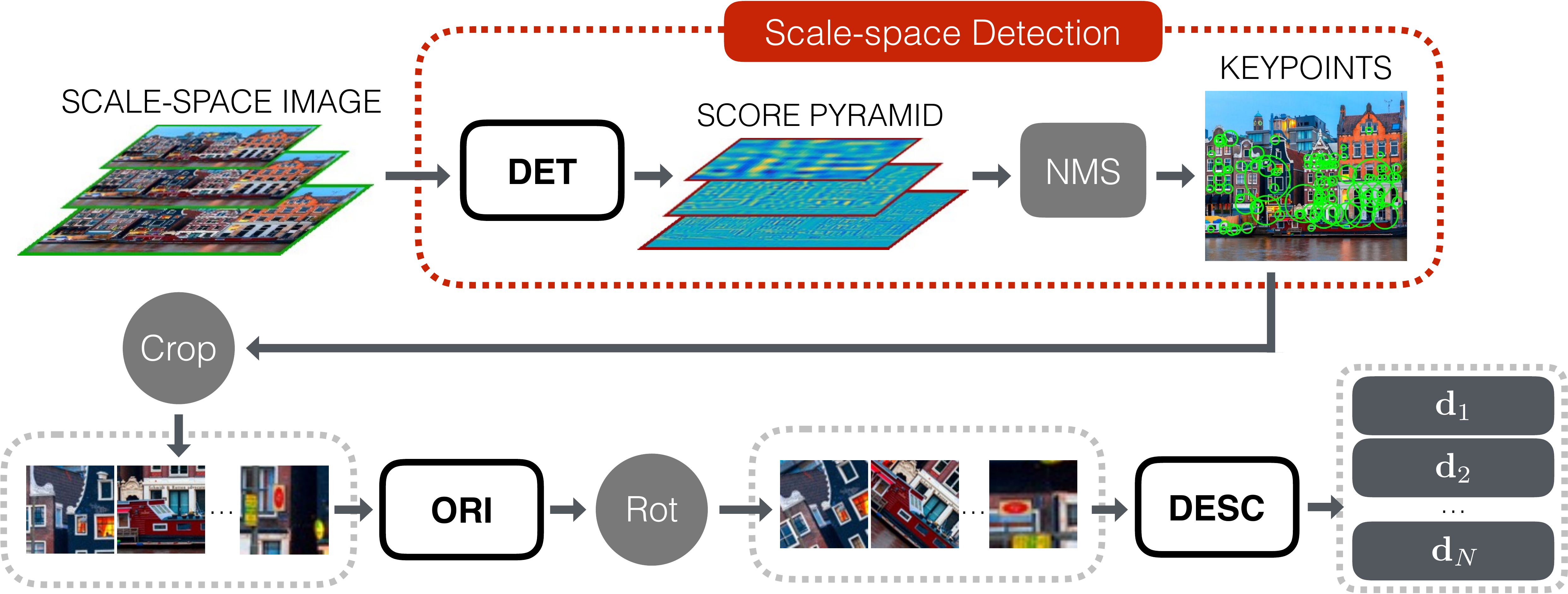}
        \caption{An overview  of our runtime architecture.   As the Orientation
          Estimator  and  the  Descriptor  only require  evaluation  at  local
          maxima,  we decouple  the Detector  and run  it in  scale space  with
          traditional  NMS  to obtain  proposals  for the  two  other
          components.}
  \label{fig:pipeline_runtime}
\end{figure}

The pipeline used  at run-time is shown  in Fig.~\ref{fig:pipeline_runtime}. As
our  method is  trained on  patches, simply  applying it  over the  image would
require the network  to be tested with  a sliding window scheme  over the whole
image.   In  practice,  this  would  be too  expensive.   Fortunately,  as  the
Orientation Estimator and  the Descriptor only need to be  run at local maxima,
we can  simply decouple  the detector  from the rest  to apply  it to  the full
image, and  replace the $\softargmax$  function by NMS,  as outlined in  red in
Fig.~\ref{fig:pipeline_runtime}.  We  then apply the Orientation  Estimator and
the Descriptor only to the patches centered on local maxima.

More exactly,  we apply the  Detector independently  to the image  at different
resolutions to obtain  score maps in scale space.  We  then apply a traditional
NMS scheme similar to that of~\cite{Lowe04} to detect feature point locations.


\section{Experimental validation}

In this section,  we first present the  datasets and metrics  we used.
 We then  present  qualitative  results, followed by a
thorough  quantitative   comparison  against   a  number   of  state-of-the-art
baselines,  which we  consistently outperform.

Finally, to better understand what elements  of our approach most contribute to
this  result, we  study  the importance  of the  pre-training  of the  Detector
component,   discussed  in   Section~\ref{subsec:detector},  and   analyze  the
performance gains attributable to each component.

\subsection{Dataset and Experimental Setup}
\label{sec:results_datasets}

We evaluate  our pipeline on three  standard datasets:
\begin{itemize}
\item The {\it Strecha} dataset~\cite{Strecha08b},  which contains 19 images of
  two scenes seen from increasingly different viewpoints.
\item The  {\it DTU}  dataset~\cite{Aanaes12}, which  contains 60  sequences of
  objects with  different viewpoints  and illumination  settings.  We  use this
  dataset to evaluate our method under viewpoint changes. 
\item The {\it Webcam} dataset~\cite{Verdie15},  which contains 710 images of 6
  scenes with strong illumination changes but seen from the same viewpoint.  We
  use this dataset to evaluate our method under natural illumination changes.
\end{itemize}

For {\it Strecha} and  {\it DTU} we use the provided  ground truth to establish
correspondences  across viewpoints.   We use  a maximum  of 1000  keypoints per
image,  and follow  the standard  evaluation protocol  of \cite{Miko03}  on the
common viewpoint region. This lets us evaluate the following metrics.
\begin{itemize}
\item Repeatability (Rep.): Repeatability of feature points, expressed
  as a  ratio.  This  metric captures the  performance of  the feature
  point detector  by reporting the  ratio of keypoints that  are found
  consistently in the shared region.
\item Nearest Neighbor mean Average Precision  (NN mAP): Area Under Curve (AUC)
  of the Precision-Recall curve, using  the Nearest Neighbor matching strategy.
  This metric captures how discriminating the descriptor is by evaluating it at
  multiple descriptor distance thresholds.
\item Matching Score (M. Score): The ratio of ground truth correspondences that
  can be recovered  by the whole pipeline over the  number of features proposed
  by the  pipeline in the  shared viewpoint  region.  This metric  measures the
  overall performance of the pipeline.
\end{itemize}

We compare  our method on  the three datasets  to the following  combination of
feature point  detectors and  descriptors, as  reported by  the authors  of the
corresponding       papers:        SIFT~\cite{Lowe04},       SURF~\cite{Bay08},
KAZE~\cite{Alcantarilla12b}, ORB~\cite{Rublee11}, Daisy~\cite{Tola10} with SIFT
detector,  sGLOH~\cite{Bellavia10} with  Harris-affine detector~\cite{Miko04c},
MROGH~\cite{Fan11}  with   Harris-affine  detector,   LIOP~\cite{Wang11e}  with
Harris-affine      detector,      BiCE~\cite{Zitnick10}     with      Edge~Foci
detector~\cite{Zitnick11},   BRISK~\cite{Leutenegger11},   FREAK~\cite{Alahi12}
with    BRISK    detector,    VGG~\cite{Simonyan14}   with    SIFT    detector,
DeepDesc~\cite{Simo-Serra15} with  SIFT detector,  PN-Net~\cite{Balntas16} with
SIFT detector, and  MatchNet~\cite{Han15} with SIFT detector.  We also consider
SIFT with Hessian-Affine keypoints~\cite{Miko02}.   For the learned descriptors
VGG,  DeepDesc, PN-Net  and MatchNet  we use  SIFT keypoints  because they  are
trained  using  a  dataset   created  with  Difference-of-Gaussians,  which  is
essentially the same  as SIFT.  In the  case of Daisy, which  was not developed
for a specific detector, we also use SIFT keypoints.
To make our results reproducible,  we provide additional implementation details
for LIFT and  the baselines in the  supplementary material.\footnote{Source and
  models will be available at \url{https://github.com/cvlab-epfl/LIFT}.}

\subsection{Qualitative Examples}

\def \figspacer {1mm}
\def \figqw {0.47}
\begin{figure}[t]
  \centering
  \includegraphics[width=\figqw\linewidth, trim = {0 30 0 0}, clip]{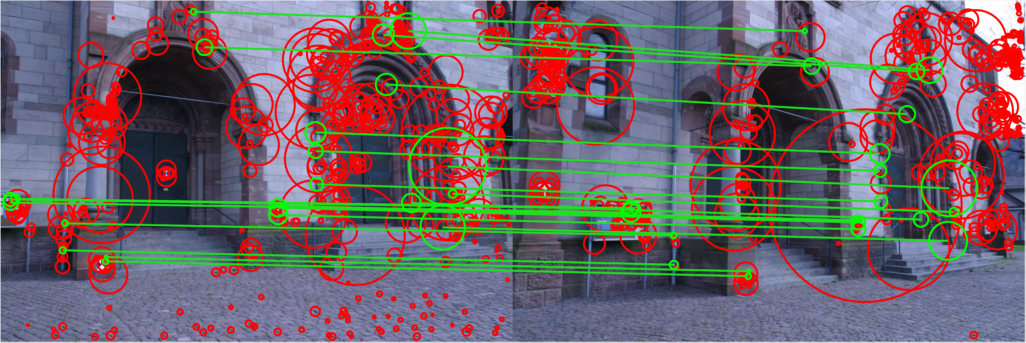}
  \includegraphics[width=\figqw\linewidth, trim = {0 30 0 0}, clip]{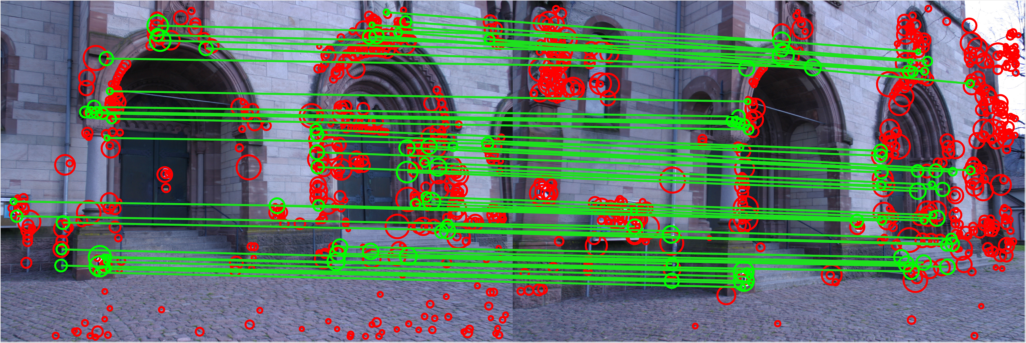}\\
  \includegraphics[width=\figqw\linewidth, trim = {0 0 0 30}, clip]{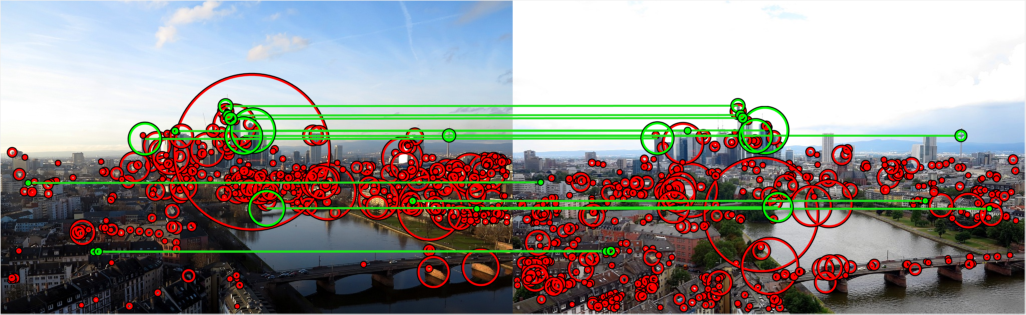}
  \includegraphics[width=\figqw\linewidth, trim = {0 0 0 30}, clip]{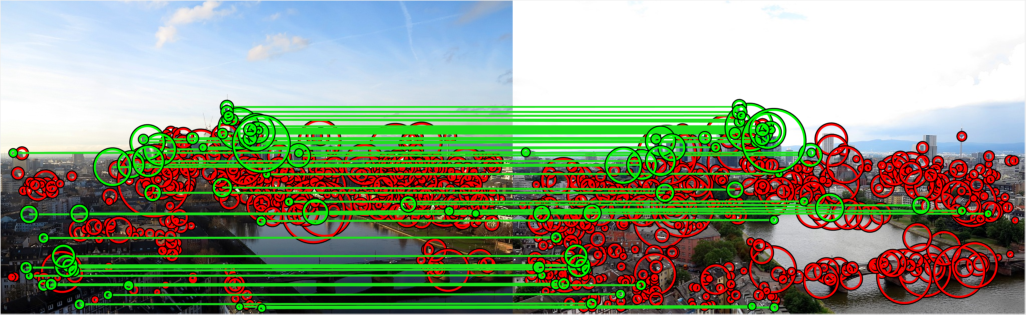}\\
  \includegraphics[width=\figqw\linewidth, trim = {0 20 0 5}, clip]{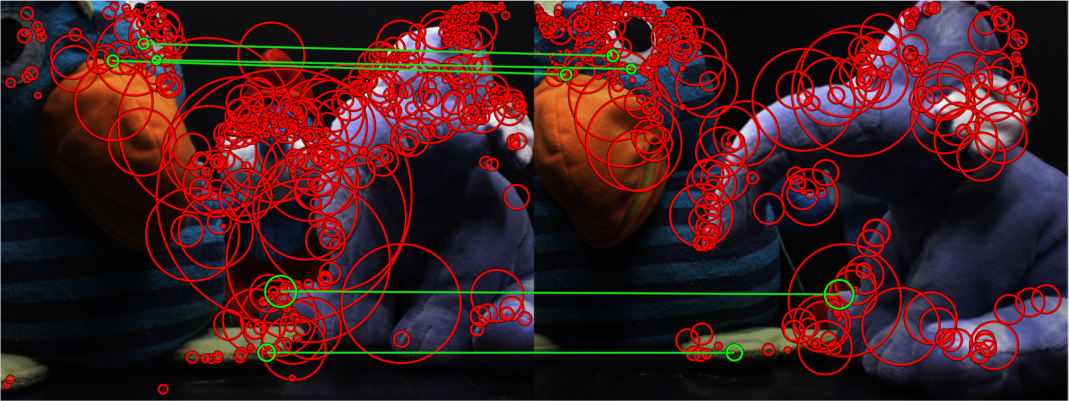}
  \includegraphics[width=\figqw\linewidth, trim = {0 20 0 5}, clip]{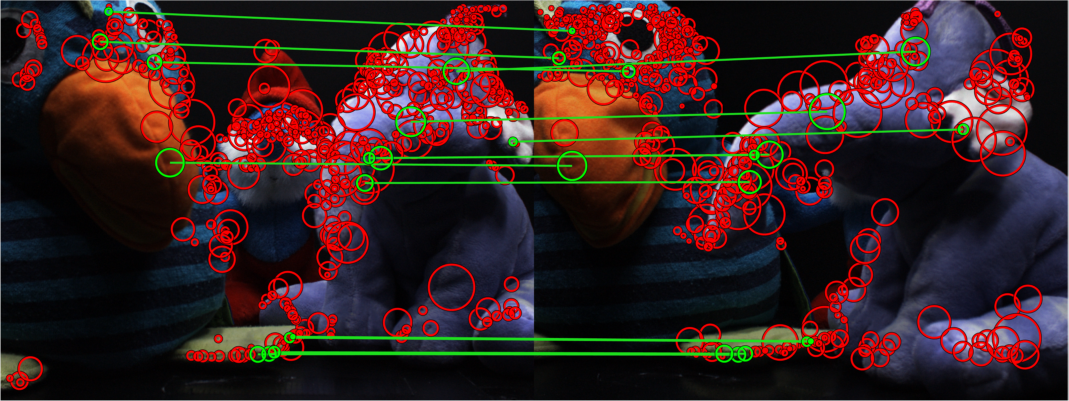}\\
  \includegraphics[width=\figqw\linewidth, trim = {0 0 0 50}, clip]{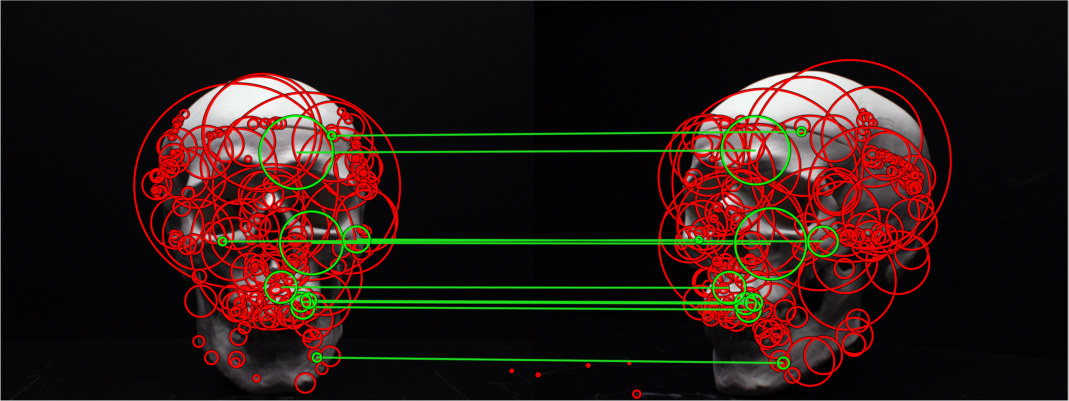}
  \includegraphics[width=\figqw\linewidth, trim = {0 0 0 50}, clip]{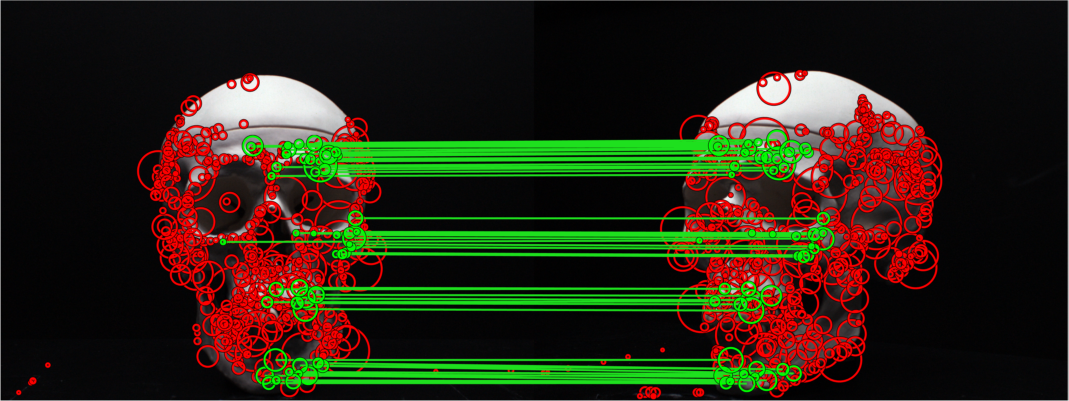}
  \caption{Qualitative local feature matching  examples of {\bf left:}
    SIFT and {\bf right:} our  method LIFT.  Correct matches recovered
    by each method are shown in green lines and the descriptor support
    regions with  red circles.  {\bf Top  row:} {\it  Herz-Jesu-P8} of
    {\it Strecha}, {\bf second row:}  {\it Frankfurt} of {\it Webcam},
    {\bf third row:} {\it Scene 7}  of {\it DTU} and {\bf bottom row:}
    {\it  Scene 19}  of  {\it  DTU}. Note  that  the  images are  very
    different from one another.}
  \label{fig:results-qualitative}
\end{figure}

Fig.~\ref{fig:results-qualitative}  shows  image   matching  results  with  500
feature  points,  for  both  SIFT  and our  LIFT  pipeline  trained  with  {\it
  Piccadilly}.  As  expected, LIFT returns more  correct correspondences across
the two images. One thing to note is  that the two DTU scenes in the bottom two
rows  are completely  different from  the  photo-tourism datasets  we used  for
training.  Given  that the  two datasets  are very  different, this  shows good
generalization properties.

\subsection{Quantitative Evaluation of the Full Pipeline}
\label{subsec:eval_full}

\begin{table}[t]
  \caption{Average matching score for all baselines.}
  \centering
  \scriptsize
  \def\arraystretch{1.1}
  \setlength{\tabcolsep}{2pt}
  \begin{tabular}{cccccccccc}
          \toprule
          & SIFT & SIFT-HesAff & SURF & ORB & Daisy & sGLOH & MROGH & LIOP & BiCE \\
          \midrule
          {\it Strecha} & .283 & .314 & .208 & .157 & .272 & .207 & .239 & .211 & .270 \\
          {\it DTU} & .272 & .274 &.244 & .127 & .262 & .187 & .223 & .189 & .242 \\
          {\it Webcam} & .128 & .164 & .117 & .120 & .120 & .113 & .125 & .086 & .166 \\
          \bottomrule
          \multicolumn{9}{c}{} \\
          & BRISK & FREAK & VGG & MatchNet & DeepDesc & PN-Net & KAZE & LIFT-pic & LIFT-rf \\
          \midrule
          {\it Strecha} & .208 & .183 & .300 & .223 & .298 & .300 & .250 & {\bf .374} & .369 \\
          {\it DTU} & .193 & .186 & .271 & .198 & .257 & .267 & .213 & {\bf .317} & .308 \\
          {\it Webcam} & .118 & .116 & .118 & .101 & .116 & .114 & .195 & .196 & {\bf .202} \\
          \bottomrule
  \end{tabular}
  \label{tbl:results-baselines}
\end{table}

\def \barw {0.30}
\begin{figure}[t]
	\centering
  \includegraphics[width=0.9\linewidth]{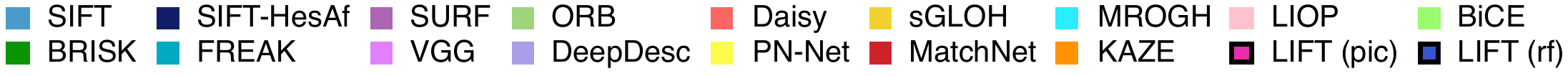}
	\includegraphics[width=\barw\linewidth]{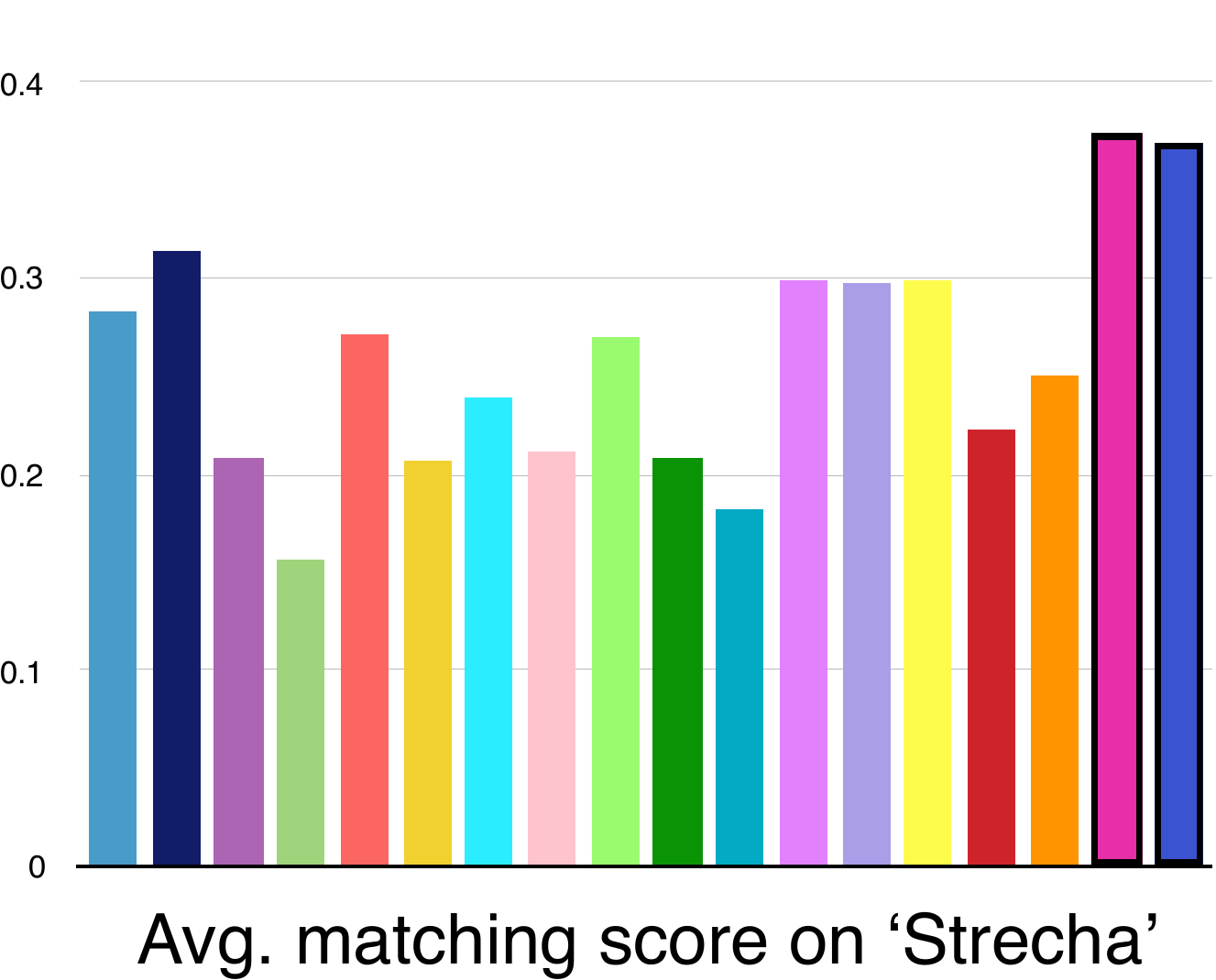}
	\includegraphics[width=\barw\linewidth]{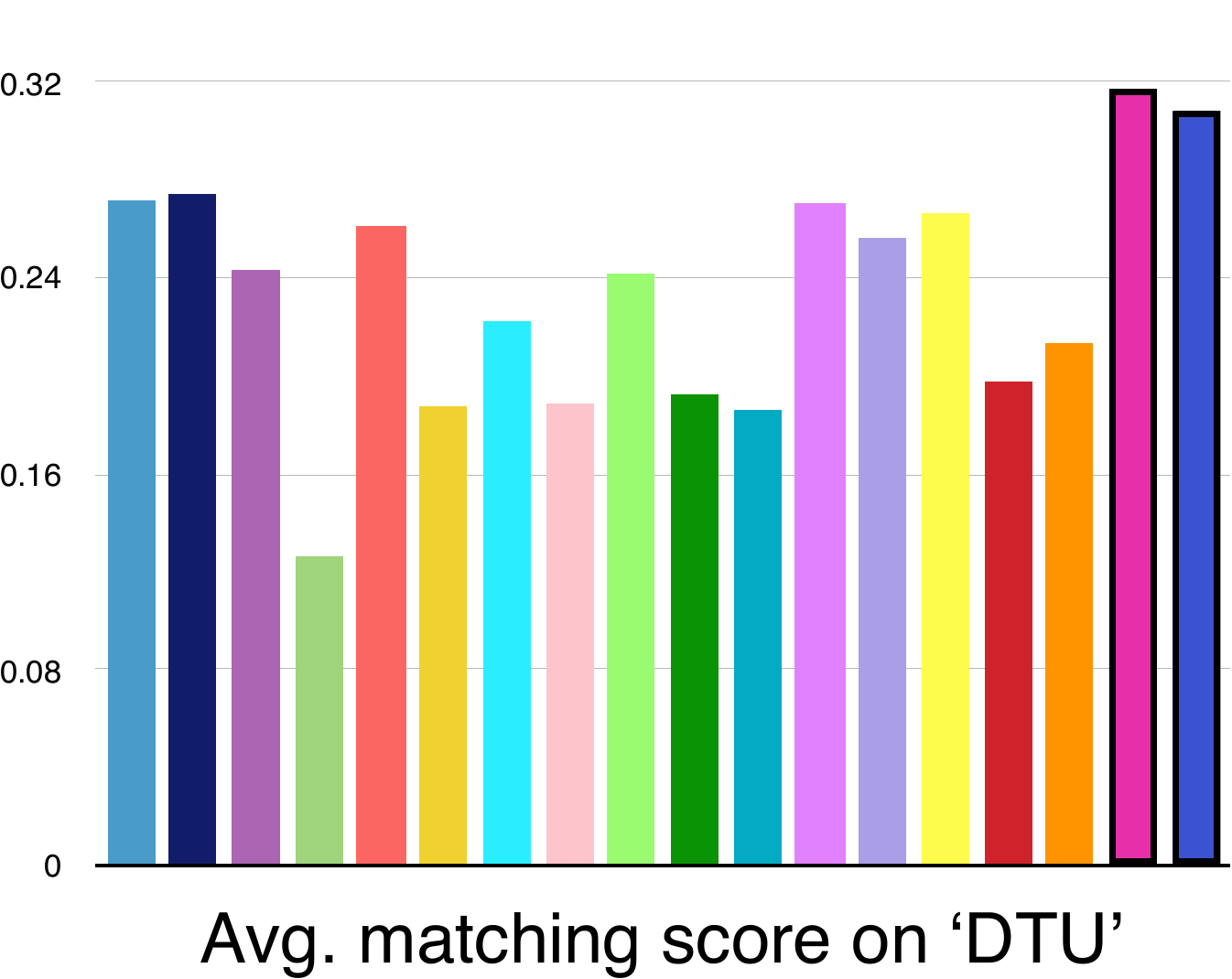}
	\includegraphics[width=\barw\linewidth]{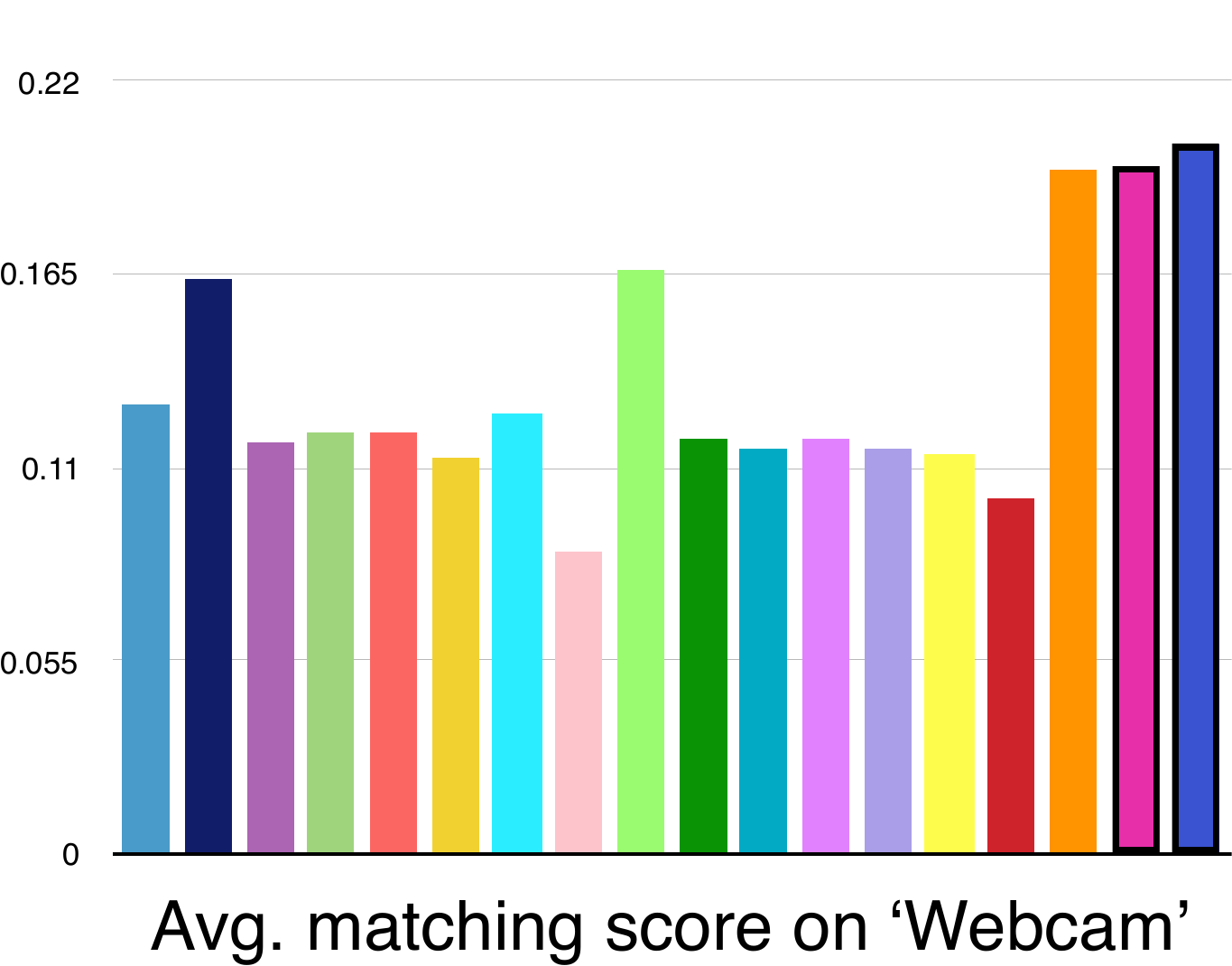}
	\caption{Average matching score for all baselines.}
	\label{fig:results-baselines}
\end{figure}

Fig.~\ref{fig:results-baselines}  shows the  average  matching  score for  all
three  datasets,  and   Table~\ref{tbl:results-baselines}  provides  the  exact
numbers for the two LIFT variants.  LIFT~(pic) is trained with {\it Piccadilly}
and LIFT~(rf) with {\it Roman-Forum}.  Both of our learned models significantly
outperform  the state-of-the-art  on {\it  Strecha} and  {\it DTU}  and achieve
state-of-the-art on {\it Webcam}.  Note that KAZE, which is the best performing
competitor on  {\it Webcam},  performs poorly  on the  other two  datasets.  As
discussed above, {\it Piccadilly} and {\it Roman-Forum} are very different from
the  datasets used  for  testing.  This  underlines  the strong  generalization
capability of our approach, which is not always in evidence with learning-based
methods.

Interestingly, on {\it DTU}, SIFT is still the best performing method among the
competitors,  even compared  to methods  that rely  on Deep  Learning, such  as
DeepDesc and  PN-Net.  Also, the gap  between SIFT and the  learning-based VGG,
DeepDesc, and PN-Net is not large for the {\it Strecha} dataset.

These results show that although a component may outperform another method when
evaluated  individually, they  may fail  to deliver  their full  potential when
integrated  into the  full pipeline,  which is  what really  matters. In  other
words, it  is important  to learn  the components  together, as  we do,  and to
consider  the  whole  pipeline  when evaluating  feature  point  detectors  and
descriptors.

\label{subsec:eval_qualitative}

\subsection{Performance of Individual Components}
\label{subsec:perf_component}

\subsubsection{Fine-tuning the Detector.}

\begin{table}[t]
  \caption{Results  on {\it  Strecha}  for  both LIFT  models  trained on  {\it
      Piccadilly} and {\it Roman-Forum}, with the pre-trained and fully-trained
    Detector.}
  \centering
  \scriptsize
  \def\arraystretch{1.}
 \setlength{\tabcolsep}{10pt}
  \begin{tabular}{ccccc}
    \toprule
                & \multicolumn{2}{c}{Trained on {\it Piccadilly}} & \multicolumn{2}{c}{Trained on {\it Roman-Forum}} \\
    \midrule
                  & Rep. & M.Score & Rep. &  M.Score \\
    Pre-trained & .436 & .367 & .447 & .368 \\
    Fully-trained & {\bf .446} & {\bf .374} & {\bf .447} & {\bf .369} \\
    \bottomrule
  \end{tabular}
  \label{tbl:pre-full}
\end{table}

\definecolor{White}{gray}{1.0}
\definecolor{Gray}{gray}{0.90}
\definecolor{LightCyan}{rgb}{0.88,1,1}
\newcolumntype{a}{>{\columncolor{Gray}}c}
\begin{table}
  \caption{Results  on {\it  Strecha}  for  both LIFT  models  trained on  {\it
      Piccadilly}  and {\it  Roman-Forum},  interchanging  our components  with
    their SIFT counterparts.}
  \centering
  \scriptsize
  \def\arraystretch{1.5}
  \setlength{\tabcolsep}{5pt}
  \begin{tabular}{cccccacca}
    \toprule
    \multicolumn{3}{c}{} & \multicolumn{3}{c}{Trained on {\it Piccadilly}} & \multicolumn{3}{c}{Trained on {\it Roman-Forum}} \\
    \rowcolor{White}
    Det. & Ori. & Desc. & Rep. & NN mAP & M.Score & Rep. & NN mAP & M.Score \\
    \midrule
    SIFT & SIFT & SIFT & \multirow{4}{*}{.428} & .517 & .282 & \multirow{4}{*}{.428} & .517 & .282 \\
    SIFT & Ours & SIFT & & .671 & .341 & & .662 & .338 \\
    SIFT & SIFT & Ours & & .568 & .290 & & .581 & .295 \\
    SIFT & Ours & Ours & & .685 & .344 & & {\bf .688} & .342 \\
    \midrule
    Ours & SIFT & SIFT & \multirow{3}{*}{\bf.446} & .540 & .325 & \multirow{3}{*}{\bf .447} & .545 & .319 \\
    Ours & Ours & SIFT & & .644 & .372 & & .630 & .360 \\
    Ours & SIFT & Ours & & .629 & .339 & & .644 & .337 \\
    \midrule
    Ours & Ours & Ours & {\bf .446} & {\bf .686} & {\bf .374} & {\bf .447}& .683 & {\bf .369} \\
    \bottomrule
  \end{tabular}
  \label{tbl:swaps}
\end{table}

Recall that we  pre-train the detector and then finalize  the training with the
Orientation    Estimator    and    the     Descriptor,    as    discussed    in
Section~\ref{subsec:detector}.  It  is therefore interesting to  see the effect
of this finalizing  stage.  In Table~\ref{tbl:pre-full} we  evaluate the entire
pipeline  with  the  pre-trained  Detector  and the  final  Detector.   As  the
pair-wise        loss        term       $\tilde{\loss}_{\text{pair}}$        of
Eq.~(\ref{eq:detector_pair_pre})  is designed  to  emulate the  behavior of  an
ideal descriptor, the pre-trained Detector already performs well.  However, the
full training pushes the performance slightly higher.

A closer look at Table~\ref{tbl:pre-full} reveals that gains are larger overall
for {\it  Piccadilly} than for {\it  Roman-Forum}. This is probably  due to the
fact that  {\it Roman-Forum} does not  have many non-feature point  regions. In
fact, the  network started to over-fit  quickly after a few  iterations on this
dataset.   The same  happened  when  we further  tried  to  fine-tune the  full
pipeline as a whole, suggesting that our learning strategy is already providing
a good global solution.

\subsubsection{Performance of individual components.}

To understand  the influence of each  component on the overall  performance, we
exchange them with their SIFT  counterparts, for both LIFT~(pic) and LIFT~(rf),
on {\it Strecha}.   We report the results in  Table~\ref{tbl:swaps}.  In short,
each time we exchange to SIFT,  we decrease performance, thus showing that each
element of  the pipeline plays and  important role.  Our Detector  gives higher
repeatability for both models.  Having better orientations also helps whichever
detector or  descriptor is being  used, and  also the Deep  Descriptors perform
better than SIFT.

One  thing  to note  is  that  our  Detector is  not  only  better in  terms  of
repeatability, but generally better in terms  of both the NN mAP, which captures
the descriptor performance, and in terms  of matching score, which evaluates the
full pipeline. This shows that our Detector  learns to find not only points that
can be found often  but also points that can be  matched easily, indicating that
training the pipeline as a whole is important for optimal performance.


\section{Conclusion}
\label{sec:conclusions}

We have  introduced a novel  Deep Network  architecture that combines  the three
components of  standard pipelines  for local  feature detection  and description
into a single differentiable network. We used Spatial Transformers together with
the $\softargmax$ function to mesh them together into a unified network that can
be  trained end-to-end  with  back-propagation. While  this  makes learning  the
network  from scratch  theoretically possible,  it  is not  {\it practical}.  We
therefore proposed an effective strategy to train it.

Our experimental  results demonstrate that our  integrated approach outperforms
the  state-of-the-art.   To further  improve  performance,  we will  look  into
strategies that allow us to take advantage even more effectively of our ability
to train the network  as a whole.  In particular, we will  look into using hard
negative mining  strategies over the whole  image~\cite{Felzenszwalb10} instead
of relying on pre-extracted patches.  This  has the potential of producing more
discriminative filters and, consequently, better descriptors.

\bibliographystyle{splncs}
\bibliography{short,vision,learning}
\end{document}